\documentclass{article}

\usepackage{microtype}
\usepackage{graphicx}
\usepackage{subfigure}
\usepackage{booktabs} 

\usepackage{hyperref}

\usepackage[accepted]{icml_fmwild}

\usepackage{amsmath}
\usepackage{amssymb}
\usepackage{mathtools}
\usepackage{amsthm}
\usepackage[utf8]{inputenc} 
\usepackage[T1]{fontenc}    
\usepackage{hyperref}       
\usepackage{url}            
\usepackage{booktabs}       
\usepackage{amsfonts}       
\usepackage{nicefrac}       
\usepackage{microtype}      
\usepackage{xcolor}         
\usepackage{graphicx} 
\usepackage{amsmath}
\usepackage{amssymb}
\definecolor{ForestGreen}{RGB}{34,139,34}
\usepackage{wrapfig,lipsum,booktabs}
\usepackage{multirow}
\usepackage{bm}
\usepackage{amsthm}
\usepackage{wrapfig,lipsum,booktabs}
\usepackage{multirow}
\usepackage{amsthm}
\usepackage{thmtools,thm-restate}
\usepackage{pythonhighlight}

\usepackage[capitalize,noabbrev]{cleveref}

\theoremstyle{plain}

\usepackage[textsize=tiny]{todonotes}

\icmltitlerunning{Sparse High Rank Adapters}

\begin{document}

\twocolumn[
\icmltitle{Rapid Switching and Multi-Adapter Fusion via Sparse High Rank Adapters}



\icmlsetsymbol{equal}{*}

\begin{icmlauthorlist}
\icmlauthor{Kartikeya Bhardwaj}{yyyy,equal}
\icmlauthor{Nilesh Prasad Pandey}{yyyy,equal}
\icmlauthor{Sweta Priyadarshi}{abcd}
\icmlauthor{Viswanath Ganapathy}{yyyy}
\icmlauthor{Rafael Esteves}{yyyy}
\icmlauthor{Shreya Kadambi}{yyyy}
\icmlauthor{Shubhankar Borse}{yyyy}
\icmlauthor{Paul Whatmough}{yyyy}
\icmlauthor{Risheek Garrepalli}{yyyy}
\icmlauthor{Mart Van Baalen}{yyyy}
\icmlauthor{Harris Teague}{yyyy}
\icmlauthor{Markus Nagel}{yyyy}
\end{icmlauthorlist}

\icmlaffiliation{yyyy}{Qualcomm AI Research. Qualcomm AI Research is an initiative of Qualcomm Technologies, Inc.}
\icmlaffiliation{abcd}{Work done while employed at Qualcomm AI Research}

\icmlcorrespondingauthor{Kartikeya Bhardwaj}{kbhardwa@qti.qualcomm.com}
\icmlcorrespondingauthor{Nilesh Prasad Pandey}{nileshpr@qti.qualcomm.com}

\icmlkeywords{Machine Learning, ICML}

\vskip 0.3in
]



\printAffiliationsAndNotice{\icmlEqualContribution} 

\begin{abstract}\vspace{-1mm}
In this paper, we propose Sparse High Rank Adapters (SHiRA) that directly finetune $1$-$2\%$ of the base model weights while leaving others unchanged, thus, resulting in a highly sparse adapter. This high sparsity incurs no inference overhead, enables rapid switching directly in the fused mode, and significantly reduces concept-loss during multi-adapter fusion. Our extensive experiments on LVMs and LLMs demonstrate that finetuning merely $1$-$2\%$ parameters in the base model is sufficient for many adapter tasks and significantly outperforms Low Rank Adaptation (LoRA). We also show that SHiRA is orthogonal to advanced LoRA methods such as DoRA and can be easily combined with existing techniques. 
\end{abstract}

\vspace{-6mm}
\section{Introduction}\label{sec::intro}
\vspace{-0.2cm}
Low Rank Adaptation (LoRA)~\cite{lora} has gained massive attention in the recent generative AI research. One of the main advantages of LoRA is its ability to be fused with pretrained models, adding no overhead during inference. Despite its success, there are still several challenges from the standpoint of deployment on the edge. First, on mobile devices, we can either avoid inference overhead in the fused mode but lose the ability to switch adapters rapidly, or suffer significant (up to 30\% higher)~\cite{hflora} inference latency while enabling \textit{rapid adapter switching} in the unfused mode (see Appendix~\ref{sec::approachMobileContd}). Second, as shown by many previous works~\cite{yu2023language, shah2023ziplora, gu2024mix}, LoRA also exhibits \textit{concept-loss} when multiple adapters are used concurrently.

Finally, recent literature also contributes important theoretical and empirical knowledge towards the value of \textit{high rank adapters}. For instance, Kalajdzievski~\cite{rslora} shows that the high rank adapters can significantly outperform low rank adapters when used with correct scaling factors. This calls for further investigation into whether other high rank adapters would outperform LoRA.
\begin{figure}[t]
  \centering
   \includegraphics[width=1.0\linewidth]{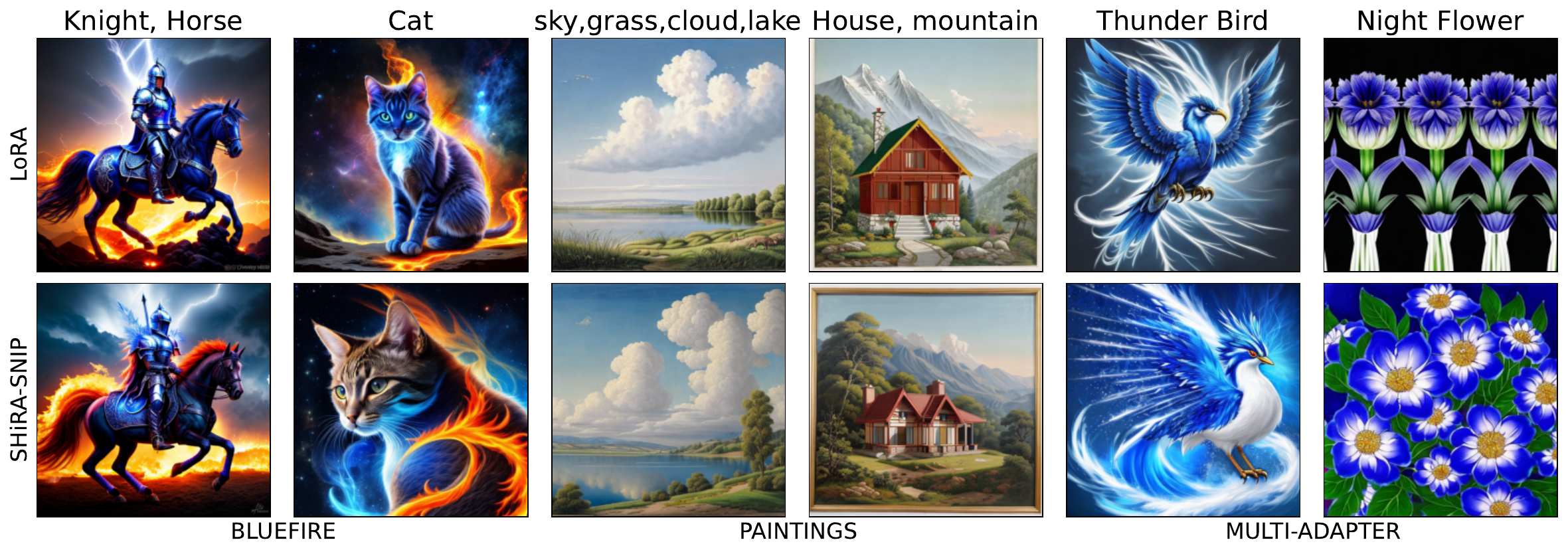}\vspace{-8mm}
   \caption{\underline{S}parse \underline{Hi}gh \underline{R}ank \underline{A}dapters (SHiRA): Changing $\sim$1-2\% weights of the pretrained model is often sufficient to achieve high performance. Due to its extreme sparsity, SHiRA enables rapid switching and also reduced concept loss during multi-adapter fusion. In contrast, LoRA modifies the majority of parameters when fused, prohibiting rapid switching on mobile devices and also experiences concept loss/artifacts during multi-adapter fusion.\vspace{-3mm}}
   \label{fig:introFigure}
\end{figure}

To address these challenges, in this paper, we propose \underline{S}parse \underline{Hi}gh \underline{R}ank \underline{A}dapters (SHiRA), a single solution to all the problems discussed above. Specifically, we make the following \textbf{key contributions}: (\textit{i})~We propose a new high rank adapter paradigm and demonstrate that changing as few as $1$-$2\%$ parameters of the original network is sufficient for many adaptation tasks. (\textit{ii})~We also conduct extensive experiments on LLaMA-7B, LLaMA2-7B, and Stable Diffusion models and demonstrate that SHiRA significantly outperforms LoRA on both single and multi-adapter fusion tasks (see Fig.~\ref{fig:introFigure}). On LLMs, we show that SHiRA achieves up to $2.7\%$ higher accuracy than LoRA on commonsense reasoning and is also complementary to more advanced variants of LoRA such as DoRA~\cite{dora}. (\textit{iii})~We further provide a latency- and memory-efficient implementation based on Parameter-Efficient Finetuning (PEFT) Library for SHiRA which trains nearly as fast as standard LoRA while consuming lower peak GPU memory (see Appendix~\ref{Memoryeff}). All our experiments can be run on a single NVIDIA-A100 GPU.

\label{sec:introduction}

\vspace{-2mm}
\section{Related Work}\label{sec::rel}

\vspace{-1mm}
\textbf{LoRA, its variants, and sparse adapters.} Many LoRA variants exist in literature: DoRA~\cite{dora}, LoRA+~\cite{hayou2024lora+}, VeRA~\cite{kopiczko2023vera}, LoRA-FA~\cite{zhang2023lora}, RS-LoRA~\cite{rslora}, among many others. The crucial difference between this literature and our work is that we develop a high rank sparse adapter which updates a fraction of weights in the pretrained weight tensor. A few other LoRA variants have also explored a combination of sparsity and low rank adaptation~\cite{nikdan2024rosa,ding2023sparse,he2022sparseadapter}. 
Most of these proposed methods promote sparsity in the adapters through either pruning~\cite{he2022sparseadapter} or adaptive rank selection~\cite{ding2023sparse}. However, since they combine their method with LoRA, the fused adapter weight still overwrites the entire pretrained weight tensor, hence prohibiting rapid adapter switching.

\textbf{Partial Finetuning.} Our work is most closely related to partial finetuning techniques proposed in the pre-LoRA era. These methods use either fixed sparse masks~\cite{sung2021training} or learned masks~\cite{zhao2020masking, guo2020parameter} to finetune a pretrained network. Note that, partial finetuning techniques have been mostly explored for relatively \textit{small} language models, and \textit{not} for recent LLMs and diffusion models. One \textit{significant} limitation of partial finetuning, as opposed to LoRA-based methods, is its high GPU memory consumption, making it impractical to be used for large generative models. Consequently, the reduced memory consumption for finetuning was a key factor to LoRA's success and its widespread adoption. To this end, we provide a memory- and latency-efficient implementation for SHiRA which trains as efficiently as LoRA, thus requiring significantly lower memory consumption compared to prior partial finetuning techniques. Further, we explore the effectiveness of sparse finetuning on both large language and vision models and provide a detailed analysis on rapid switching and multi-adapter fusion of the high rank adapters.

\textbf{Multi-Adapter Fusion.} Existing methods like~\cite{gu2024mix, yu2023language, shah2023ziplora} use the base LoRA with non-trivial postprocessing~\cite{yu2023language, shah2023ziplora}. In contrast, we introduce a new adapter where multiple concepts naturally do not interfere with each other. Our work is also orthogonal to the prior multi-adapter fusion work and can be combined with these techniques.

\label{sec:Related_Work}
\vspace{-0.1cm}
\section{Proposed Approach}\label{sec::approach}

\subsection{\underline{S}parse \underline{Hi}gh \underline{R}ank \underline{A}dapters (SHiRA)} \label{shira-approach}
\begin{figure}[t]
  \centering
   \includegraphics[width=1.0\linewidth]{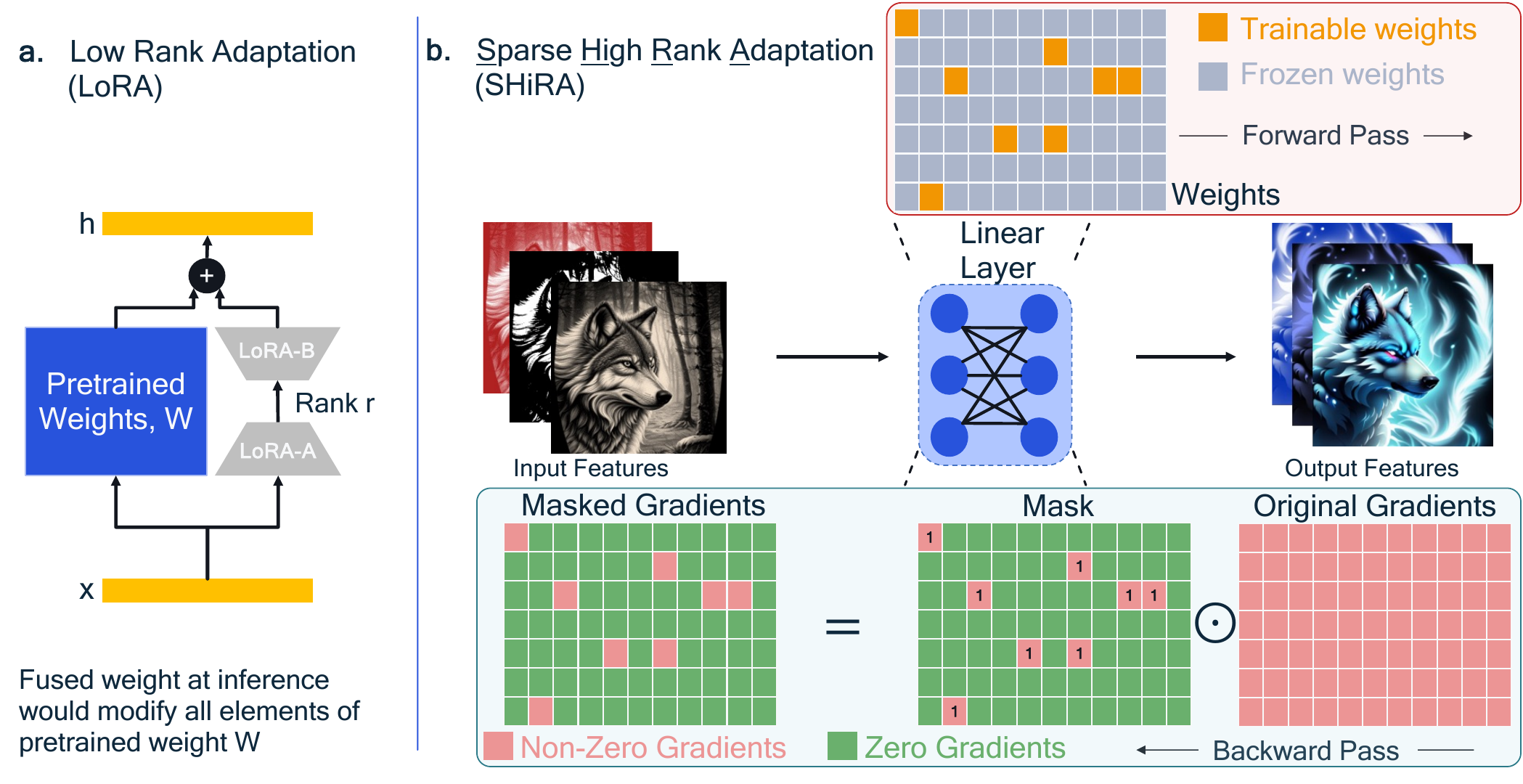}\vspace{-8mm}
   \caption{(a) LoRA appends two low rank weights that can be fused into the base weights at inference. However, this modifies all weights and prevents rapid switching. (b) SHiRA finetunes very few pretrained weights by exploiting gradient-masking during training. We show that finetuning as low as $1$-$2\%$ parameters is sufficient to achieve high accuracy on many adapter tasks.}
   \label{fig:shiraApproach}
   \vspace{-0.2cm}
\end{figure}

SHiRA exploits highly sparse trainable parameters in the pretrained base model. Specifically, we do not add any new weights to the forward pass like LoRA (see Fig.~\ref{fig:shiraApproach}(a)) but rather make a small percentage of existing weights trainable (see Fig.~\ref{fig:shiraApproach}(b) top). To this end, we first create an extremely sparse ($\sim$$98$-$99\%$ zeros) mask $\mathcal{M}\in\mathbb{R}^{n\times m}=\{0,1\}^{n\times m}$, where $n, m$ are dimensions of the pretrained weight matrix. $\mathcal{M}$ is then used to mask the gradients during backpropagation using a Hadamard product (see Fig.~\ref{fig:shiraApproach}(b) bottom). As a result, very few parameters get updated during training and our adapter consists of just those sparse weights. 
Various strategies to create the mask $\mathcal{M}$ are given below:\vspace{-4mm}
\begin{enumerate}\setlength{\itemsep}{-0.4em}
    \item \textbf{SHiRA-Struct:} In this structured mask, only certain rows or columns of the weight as well as its diagonal are set to be trainable. The diagonal makes the mask high rank whereas the structured trainable rows/ columns lead to a rank 1 adapter, making the mask a combination of a sparse high rank and a rank 1 adapter.
    \item \textbf{SHiRA-Rand:} This mask is obtained by randomly setting $1$-$2\%$ parameters trainable. 
    \item \textbf{SHiRA-WM:} Top-K parameters are trained based on their absolute weight magnitudes (WM) for each layer.
    \item \textbf{SHiRA-Grad:} Top $1$-$2\%$ weights that receive the highest gradient magnitudes based on a calibration set are trained for each layer.
    \item \textbf{SHiRA-SNIP:} This mask is based on SNIP metric used in the pruning literature~\cite{lee2018snip}. SNIP combines weight magnitude and gradient strategies, i.e., gradient magnitude times the weight magnitude. 
\end{enumerate}\vspace{-1mm}

\vspace{-0.3cm}
\subsection{Rapid Switching and Multi-Adapter Fusion}
\begin{figure}[t]
  \centering
   \includegraphics[width=1.0\linewidth]{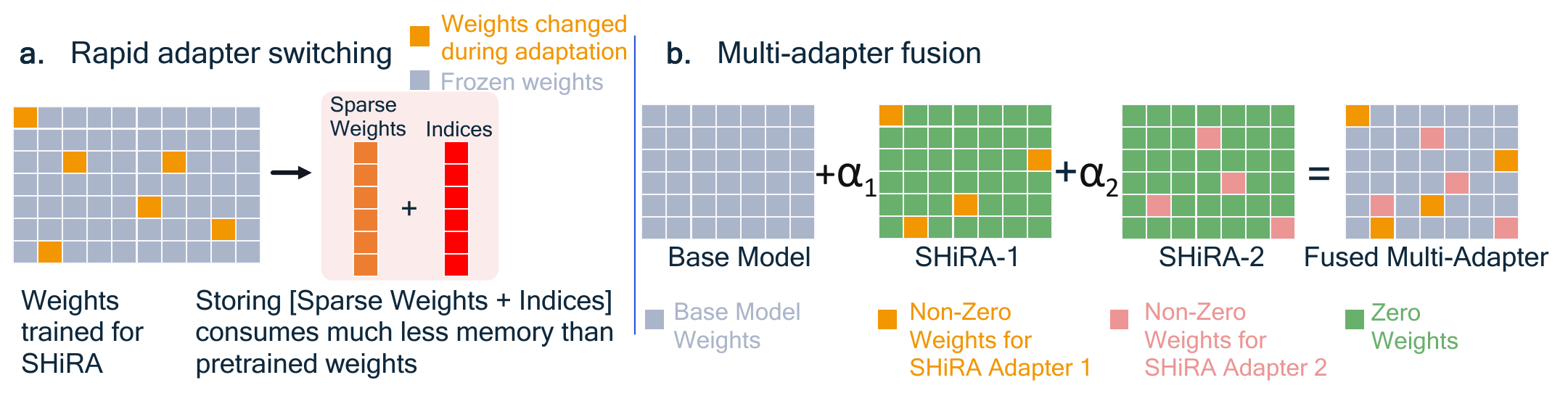}\vspace{-8mm}
   \caption{(a) Rapid switching: SHiRA adapters can be stored as sparse weights and their indices which can be loaded on the base model. Since only $1$-$2\%$ weights need to be overwritten, the adapter can be efficiently switched with different weights at inference, eliminating the need for a separate fusion stage. (b) Multi-adapter fusion: Multiple adapters can be fused together by naively adding them together and then loading the resulting sparse weights.}
   \label{fig:shiraBenefits}
   \vspace{-0.1cm}
\end{figure}

\vspace{-0.2cm}
Since very few base weights change during the SHiRA training, we can simply extract them out and store them as sparse weights and their indices (see Fig.~\ref{fig:shiraBenefits}(a)). Hence, SHiRA is comparable to LoRA in model size but overwrites only a fraction of the pretrained weights at inference time. In contrast, LoRA fuses into base weights as $W_{new} = W + AB$ and changes the entire weight. Note that, we do not actually need to fuse SHiRA but rather just need to overwrite the modified value at the correct index in the pretrained weight tensor. This enables rapid switching on resource-constrained devices. To verify that SHiRA indeed provides rapid switching benefits compared to LoRA, we provide an optimized implementation based on \texttt{scatter\_op} to overwrite base model weights instead of fusing them like LoRA. We demonstrate that on a CPU, \textbf{weight loading for SHiRA adapters can be up to }$\bm{10\times}$ \textbf{faster than equivalent LoRA fusing} (see Appendix~\ref{sec::scatterop} and Fig~\ref{fig:speed}).

Next, we discuss multi-adapter fusion in SHiRA. Suppose, we are given two adapters $\mathcal{A}_1$ and $\mathcal{A}_2$ that represent two separate concepts. Now, consider the product $\mathcal{A}_1^T\mathcal{A}_2$ to evaluate the relative orthogonality of two adapters. Note that, for SHiRA, the adapters are based on sparse masks $\mathcal{M}_{1}$ and $\mathcal{M}_{2}$ which are both up to 98-99\% sparse. Due to this high sparsity of SHiRA masks, $\mathcal{A}_1^T\mathcal{A}_2$ contains a much higher number of zeros compared to equivalent dense LoRA adapters (since fused LoRA adapter $AB$ is a dense matrix). Therefore, we hypothesize that this high sparsity property of SHiRA adapters would result in significantly lower interference between adapters and, hence, reduced concept loss. We empirically validate this in section~\ref{sec::exp}.

\vspace{-0.2cm}
\subsection{Memory- and Latency-Efficient SHiRA Training.}\vspace{-0.2cm}
We can have two implementations: (\textit{i})~backward hook based gradient masking to turn any trainer into SHiRA training (refer Appendix~\ref{Latencyeff}) (\textit{ii})~PEFT based implementation. As discussed in Appendix~\ref{Memoryeff}, the PEFT based SHiRA implementation consumes \textbf{\bm{$16.63\%$} lower peak GPU memory and trains almost at a similar speed as LoRA}. On the contrary, DoRA exhibits a $40.99\%$ and $28.9\%$ increase in memory and training time, respectively, compared to LoRA.

\label{sec:method}
\vspace{-3mm}
\section{Experiments}\label{sec::exp}
\vspace{-2mm}
\subsection{Training Setup and Datasets}
\vspace{-3mm}

\begin{table}[!t]
    \centering
    \fontsize{7.0pt}{5.75pt}\selectfont
    \begin{tabular} {  lcccccccc }
    \toprule
    
    \multirow{1}{*}{\textbf{Style}} & \multirow{1}{*}{\textbf{Method}}  & \textbf{\%Params} & \multicolumn{2}{c}{\textbf{HPSv2 score($\uparrow$)}} \\

    & & & $\alpha=1$ & $\alpha=0.5$ \\
    
    \hline
    \midrule

    \multirow{6}{*}{Paintings} & LoRA & 3.84 &$24.7 \pm1.8$ & $31.3 \pm 1.5$ \\
    & SHiRA-Struct &  \textbf{1.99} &$\bm{31.2 \pm 1.7}$ &  $\bm{33.0 \pm 1.8}$ \\
    & SHiRA-Rand & 2.05&$30.7 \pm 1.9$ &  $32.7 \pm 1.9$ \\
    & SHiRA-WM & 2.05&$29.7 \pm 1.9$ & $32.1 \pm 1.8$ \\
    & SHiRA-Grad & 2.05&$30.3 \pm 1.8$  & $32.3 \pm 1.8$ \\
    & SHiRA-SNIP & 2.05&$29.8 \pm 1.8$ & $31.6 \pm 1.8$ \\
    \midrule
    \multirow{6}{*}{Bluefire} & LoRA & 3.84&$32.6 \pm 1.9$ &  $33.6 \pm 1.6$ \\
    & SHiRA-Struct & \textbf{1.99} &$\bm{34.2 \pm 1.6}$  & $\bm{34.1 \pm 1.5}$  \\
    & SHiRA-Rand & 2.05&$33.4 \pm 1.9$  & $33.7 \pm 1.7$ \\
    & SHiRA-WM & 2.05&$31.9 \pm 2.1$  & $33.1 \pm 1.7$ \\
    & SHiRA-Grad & 2.05&$\bm{34.2 \pm 1.5}$ & $33.7 \pm 1.7$ \\
    & SHiRA-SNIP & 2.05&$33.7 \pm 1.7$  & $33.7 \pm 1.6$ \\

    \bottomrule
\end{tabular}\vspace{-3mm}
    \caption{Comparison between LoRA and SHiRA schemes with respect to HPSv2 metric.}\vspace{-4mm}
    \label{tab:t2i_table}
\end{table}

\begin{table*}[t]
    \centering
    \scalebox{0.9}{
    \addtolength{\tabcolsep}{-2pt}
    \fontsize{7.0pt}{5.75pt}\selectfont
    \begin{tabular} {  lccccccccccc }
    \toprule
    \textbf{Model} & \textbf{\%Params} & \textbf{\%C} & \textbf{BoolQ($\uparrow$)} & \textbf{PIQA($\uparrow$)} & \textbf{Arc-e($\uparrow$)} & \textbf{Arc-c($\uparrow$)} & \textbf{WG($\uparrow$)} & \textbf{OBQA($\uparrow$)} & \textbf{HS($\uparrow$)} &  \textbf{SIQA($\uparrow$)} &\textbf{Avg.($\uparrow$)} \\\hline
    \midrule
    LoRA & \textbf{0.83} & {\color{red} 66.72} & 68.9 & 80.7 & 77.8 & 61.3 & 78.8 & 74.8 & 78.1 & 77.4 & 74.7 ({\color{red} +0\%}) \\
    \midrule
    SHiRA-Grad  & 1.0 & {\color{ForestGreen}\textbf{1.0}} & 68.4 & 80.9 & 80.2 & 64.7 & \textbf{80.4} & 78.2 & 80.3 & \textbf{79.4} & 76.6 ({\color{ForestGreen} +1.9\%})\\
    \midrule
     SHiRA-WM  & 1.0 & {\color{ForestGreen}\textbf{1.0}} & \textbf{69.6} & \textbf{81.6} & \textbf{81.5} & 66.5 & 79.8 & 79.4 & 79.6 & 77.8 & 77.0 ({\color{ForestGreen} +2.3\%})\\
    \midrule
    \textbf{SHiRA-SNIP}  & 1.0 & {\color{ForestGreen}\textbf{1.0}} & 68.3 & 80.6 & \textbf{81.5} & \textbf{67.9} & 80.0 & \textbf{79.6} & \textbf{82.1} & 79.1 & \textbf{77.4} ({\color{ForestGreen} +2.7\%})\\
     \hline 
    \midrule
    DoRA & 0.84 & {\color{red} 66.72} & 68.5 & \textbf{82.9} & 81.4 & \textbf{65.8} & \textbf{80.8} & \textbf{81.0} & \textbf{84.8} & \textbf{79.6} & \textbf{78.1} ({\color{ForestGreen} +0\%})\\
    \midrule
    SHiRA-WM-DoRA & 6.25$^*$ & {\color{ForestGreen}\textbf{1.0}} & \textbf{70.9} & 81.9 & \textbf{81.7} & 64.9 & \textbf{80.8} & 79.2 & 84.5 & 78.6 & \textbf{77.8} ({\color{red} -0.3\%})\\
    \bottomrule
\end{tabular}
}\vspace{-3mm}
    \caption{Evaluation of LLaMA-7B models on Commonsense Reasoning. WG and HS denote WinoGrande and HellaSwag, respectively. \%C represents parameters changed in the fused mode.  ($\uparrow$): the higher the better. {\color{ForestGreen} Green} denotes improvement over baselines. $^*$Trained by masking a high-rank DoRA with a WM mask of top 1\% weights, thus changing only 1\% of the model during both training and inference.} 
    \label{tab: CommonSense-Single-Adapter}
\end{table*}

\begin{table*}[t]
\centering
\scalebox{0.925}{
    \addtolength{\tabcolsep}{-2pt}
    \fontsize{7.0pt}{5.75pt}\selectfont
    \begin{tabular} {  lccccccccccc }
    \toprule
    \textbf{Model} & \textbf{\%Params} & \textbf{\%C} & \textbf{BoolQ($\uparrow$)} & \textbf{PIQA($\uparrow$)} & \textbf{Arc-e($\uparrow$)} & \textbf{Arc-c($\uparrow$)} & \textbf{WG($\uparrow$)} & \textbf{OBQA($\uparrow$)} & \textbf{HS($\uparrow$)} &  \textbf{SIQA($\uparrow$)} &\textbf{Avg.($\uparrow$)} \\\hline
    \midrule
    LoRA & \textbf{0.83} & {\color{red}66.72} & 69.90 & 79.9 & 79.8 & 64.7 & 82.6 & 81.0 & 83.6 & \textbf{79.5} & 77.61 ({\color{red} +0\%})\\
    \midrule
    DoRA & 0.84 & {\color{red}66.72} & \textbf{71.8} & \textbf{83.7} & \textbf{83.7} & 68.2 & \textbf{82.6} & \textbf{82.4} & 89.1 & 76.0 & \textbf{79.68} ({\color{ForestGreen} +2.07\%})\\
    \midrule
    \textbf{SHiRA-SNIP}  & 1.0 & {\color{ForestGreen}\textbf{1.0}} & 70.42 & 81.71 & 83.25 &  \textbf{68.6} & 80.51 & 81.0 &  \textbf{89.78} & 79.01 & \textbf{79.28} ({\color{ForestGreen} +1.67\%}) \\
    
    \midrule
\end{tabular}
}\vspace{-3mm}
    \caption{Results for LLaMA2-7B on Commonsense Reasoning.}
    \label{tab: CommonSense-2}
\end{table*}

\begin{table*}[t!]
    \addtolength{\tabcolsep}{-2pt}
    \centering
\scalebox{0.93}{
    \fontsize{7.0pt}{5.75pt}\selectfont
    \begin{tabular} {  lcccccccccc }
    \toprule
    \multicolumn{1}{c}{} & \multicolumn{4}{c}{\textbf{Single Adapter}} & & \multicolumn{4}{c}{\textbf{Multi-Adapter}} &\\
    \cmidrule{2-5} \cmidrule{7-10}
    \textbf{Model} & \textbf{BoolQ($\uparrow$)} & \textbf{PIQA($\uparrow$)} & \textbf{Arc\_e($\uparrow$)} & \textbf{Avg($\uparrow$)}&  & \textbf{BoolQ($\uparrow$)} & \textbf{PIQA($\uparrow$)} & \textbf{Arc\_e($\uparrow$)}& \textbf{Avg($\uparrow$)} & \textbf{\%Drop ($\downarrow$)}\\\hline
    \midrule
    LoRA & \textbf{80.52} & 79.05 & 75.67 & 78.41 & & 77.22 & 71.27 & 57.45 & 67.33 ({\color{red} +0\%}) & {\color{red} 11.08} \\
    \midrule
    SHiRA-WM & 78.07 & \textbf{79.71} & \textbf{77.57} & \textbf{78.45}& & \textbf{77.43} & \textbf{76.88} & \textbf{67.76} & \textbf{74.02} ({\color{ForestGreen} +6.69\%}) & {\color{ForestGreen} 4.43}\\
    \bottomrule
\end{tabular}
}\vspace{-3mm}
    \caption{Multi-adapter fusion of independently trained SHiRA and LoRA adapters on  BoolQ, PIQA, and Arc-Easy for LLaMA2-7B. \%Drop is calculated as drop in average accuracy for multi-adapter fusion compared to the single adapter average accuracy for each adapter.} 
    \label{tab: multishira}
\vspace{-0.3cm}
\end{table*}

For the vision tasks, we use the RealisticVision-v3 model checkpoint for Stable Diffusion-v1.5, and finetune it using different adapters on two style transfer datasets collected using public domain images. The first dataset is called Bluefire which provides a ``blue fire'' effect to images. The second dataset is a painting dataset, which gives a ``paintings'' effect (see Appendix section \ref{dataset-desp} for more details). For both these datasets, we conduct single adapter and multi-adapter experiments and quantify the image quality using Human Preference Score-V2 (HPSv2)~\cite{wu2023human}. 

On the language domain, we experiment with LLaMA 7B~\cite{llama}, LLaMA2-7B~\cite{llama2} and evaluate it on various commonsense reasoning tasks such as HellaSwag, PIQA, SIQA, BoolQ, Arc-easy, Arc-challenge, OpenBookQA and Winogrande. Specifically, we follow the setup adopted by \cite{llmadapters, dora} for training and evaluating LoRA~\cite{lora}, DoRA~\cite{dora}, and SHiRA on downstream tasks. Also, similar to our vision investigations, we conduct single and multi-adapter experiments on LLMs as well.

\subsection{Vision Results}

\begin{figure}[t]
  \centering
   \includegraphics[width=1.0\linewidth]{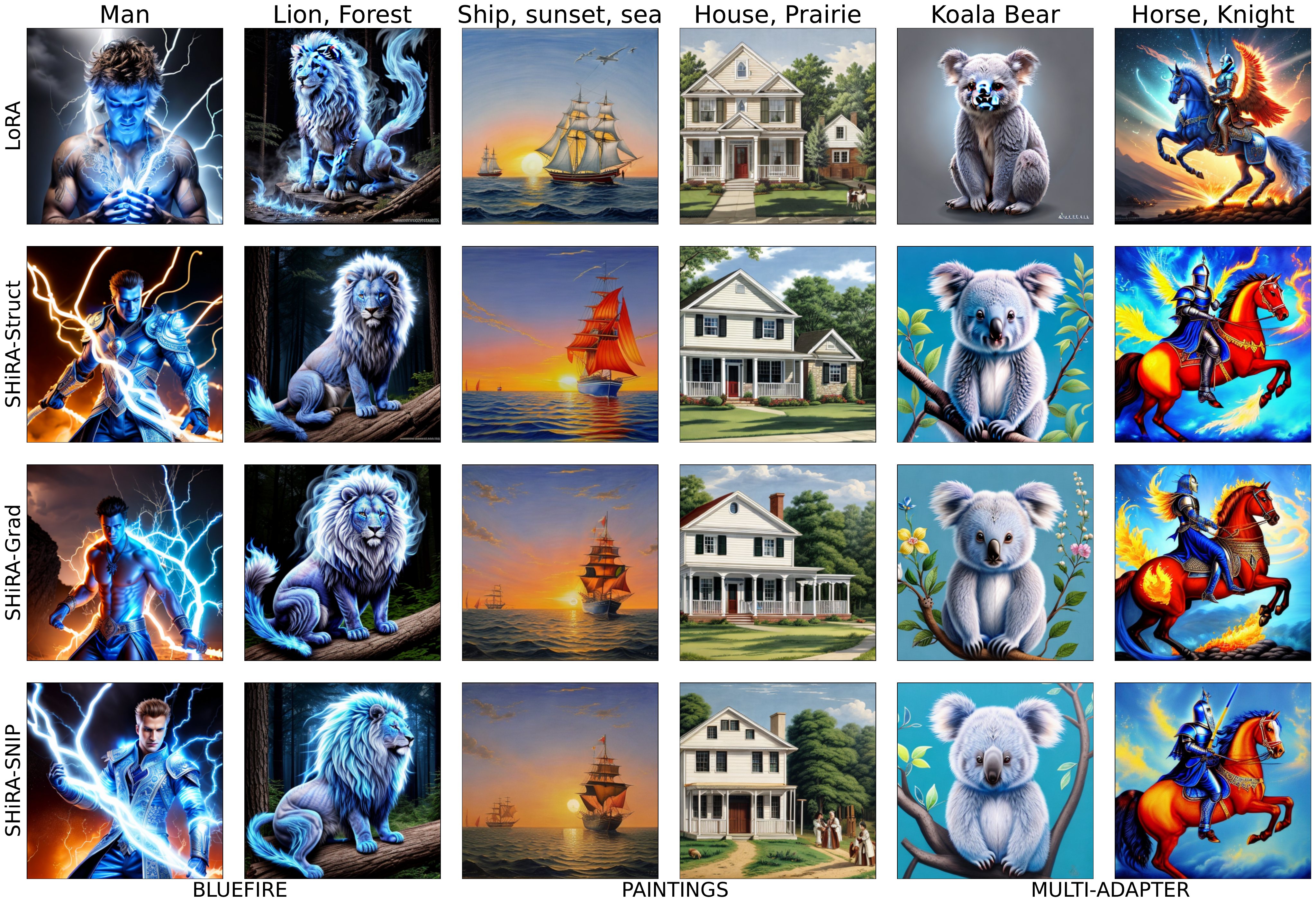}\vspace{-6mm}
   \caption{Comparison between different SHiRA masking methods for single and multi adapter image generation. For multi-adapter fusion, SHiRA-Struct outperforms all other adapters. SHiRA does not have artifacts and concept-loss like LoRA (see Koala/Knight).}
   \label{fig:shiraFigure}\vspace{- 0.5 em}
\end{figure}
\subsubsection{Impact of Various SHiRA Masks}\label{single-shira-lvm}
\vspace{-0.1cm}
We first evaluate the image quality for SHiRA and LoRA on Paintings and Bluefire datasets for both single and multi-adapter usecases. Fig.~\ref{fig:introFigure} demonstrates comparison between SHiRA-SNIP and LoRA. As evident, by merely changing 1-2\% pretrained weights, SHiRA generates high quality images for both finetuning tasks. 

Next, we compare various types of SHiRA masks in Fig.~\ref{fig:shiraFigure}. Clearly, all SHiRA schemes produce impressive images for different prompts and significantly outperform LoRA. We further quantify the image quality using HPSv2 for each of the masks. The results are presented in Table~\ref{tab:t2i_table}. As evident, all variants of SHiRA consistently achieve superior or similar HPSv2 scores than LoRA, especially for larger $\alpha$, the scaling factor used for modifying the intensity of the adapter during inference (see Appendix~\ref{alpha-effect}).

\vspace{-1mm}
\subsubsection{SHiRA Improves Multi-Adapter Fusion}\label{lvm-2}
We validate the effectiveness of various SHiRA schemes for multi-adapter fusion. In Fig.~\ref{fig:introFigure} and \ref{fig:shiraFigure}, both SHiRA-Struct and SNIP (right two columns) are clearly better at capturing both concepts than LoRA. For example, the knight image in Fig.~\ref{fig:shiraFigure} generated with LoRA seems to lose most of the paintings concept (more results in Fig.~\ref{fig:shiraFigure4}, Appendix~\ref{AppMoreRes}). 

\vspace{-1mm}
\subsection{Language Results}

\subsubsection{Single Adapter SHiRA Finetuning}\label{single-adapter-llm}
\vspace{-0.1cm}
For LLMs, different SHiRA adapters are trained on the combined 170K sample commonsense reasoning dataset released by~\cite{llmadapters, dora}. Similar to~\cite{dora}, we train our SHiRA adapters for 3 epochs and compare it against the LoRA baselines. As shown in Table~\ref{tab: CommonSense-Single-Adapter} for LLaMA-7B, various SHiRA adapters outperform LoRA by 1.9-2.7\% on an average on LLaMA-7B. Importantly, SHiRA only modifies 1\% base parameter weights as compared to \textbf{66.72\%} (\textbf{4.5B weights}) changed by LoRA in the fused mode, thus enabling rapid adapter switching on edge devices. Of note, since SHiRA-Struct is a combination of rank-1 adapter and a sparse diagonal adapter, it did not perform well on more complex LLM tasks.

SHiRA is orthogonal to newer adapters like DoRA and can be easily integrated with them. To demonstrate this, we create a high rank weight decomposed adapter similar to DoRA and then mask $99\%$ of this weight using SHiRA. As shown in Table~\ref{tab: CommonSense-Single-Adapter}, our proposed adapter benefits from DoRA based finetuning and achieves almost comparable performance (within 0.3\%) to DoRA on an average, with an added benefit of changing only 1\% parameters at inference time.  Finally, we experiment with LLaMA2-7B and demonstrate that SHiRA-SNIP -- which achieved the best results on LLaMA-7B -- yields significant gains compared to LoRA and nearly the same accuracy as DoRA (within 0.4\%, see Table~\ref{tab: CommonSense-2}).

\subsubsection{Multi-Adapter Fusion on LLMs}
We now extend our analysis to the multi-adapter fusion setting. To this end, we create a \textit{new} setup where we independently train multiple adapters on training sets of individual commonsense reasoning benchmarks, i.e., one adapter each for BoolQ, PIQA, and Arc-Easy. In contrast, each adapter in section~\ref{single-adapter-llm} was trained on a combined dataset containing 170K samples from all eight commonsense benchmarks, as proposed in~\cite{llmadapters, dora}. In the present section, the goal is to evaluate how much accuracy drop various adapters experience after multi-adapter fusion. Due to its simplicity towards constructing a mask, we use SHiRA-WM which consists of top 1\% parameters being trained for all tasks. As shown in table~\ref{tab: multishira}, multi-SHiRA-WM outperforms multi-LoRA on all the three benchmarks, outperforming LoRA by  6.69\% accuracy on average.

\vspace{-1mm}
\vspace{-0.2cm}
\label{sec:experiments}
\section{Conclusion}\label{sec::conclusion}
\vspace{-1.1mm}
In this paper, we have proposed SHiRA, a new high rank adapter paradigm to demonstrate that even finetuning merely 1-2\% parameters of the pretrained generative models is sufficient to achieve high performance on many adapter tasks. We demonstrate SHiRA's ability to rapidly switch adapters and to avoid concept loss with naive multi-adapter fusion, contrary to LoRA which suffers from both of these problems. We have further conducted extensive single- and multi-adapter experiments on several vision and language tasks and demonstrate the effectiveness of SHiRA compared to LoRA. Our latency- and memory-efficient PEFT-based implementation for training SHiRA runs at nearly the same speed as LoRA while consuming about 16\% lower peak GPU memory. Finally, for inference, we have provided a \texttt{scatter\_op} based method that can load our SHiRA up to $10\times$ faster than equivalent LoRA fusion on a CPU, thus demonstrating our rapid switching benefits. 

For further reading, please refer to our extended work~\cite{bhardwaj2024sparse}. 
\label{sec:conclusion}


\bibliography{example_paper}
\bibliographystyle{icml2024}



\newpage
\onecolumn
\appendix
\section{Edge Deployment Challenges for LoRA}\label{sec::approachMobileContd}
There are three existing deployment options for LoRA: (\textit{i})~fuse the adapter offline and then deploy on-device: this 
changes a large fraction of the weight tensors compared to base model which prohibits rapid adapter switching since it will increase DRAM traffic considerably; (\textit{ii})~keep the adapter unfused and run the adapter in unfused mode: this can help with rapid adapter switching but would incur  significant additional latency (up to 30\% higher) as shown in~\cite{hflora}, since we would have LoRA branches in the forward pass during inference; (\textit{iii})~use the Huggingface/Diffusers pipeline~\cite{hflora} which is used for server-grade GPUs. This pipeline consists of \texttt{load}$\rightarrow$\texttt{fuse}$\rightarrow$\texttt{inference}$\rightarrow$\texttt{unfuse}$\rightarrow$\texttt{unload} to switch adapters. In the last option, unfused LoRA-A and LoRA-B weights (see Fig.~\ref{fig:shiraApproach}(a)) are first loaded into the memory and then fused into the base model by computing $W_{new} = W+AB$; this new weight is used for inference. To switch the adapter, we can unfuse the adapter as $W=W_{new}-AB$ and then unload existing LoRA weights to load the new ones. 

To understand the overhead of each of the stages to the standard huggingface LoRA inference pipeline (i.e., \texttt{load}, \texttt{fuse}, \texttt{unfuse}, \texttt{unload}), we experiment with the pipeline provided in~\cite{hflora} and iteratively add adapters to SDXL model~\cite{sdxl}. As evident 
\begin{wraptable}{r}{6.3cm}
    \addtolength{\tabcolsep}{-2pt}
    \centering
    \fontsize{6.75pt}{6.75pt}\selectfont
    
    \begin{tabular} {  lccc }
    \toprule
    \centering \textbf{Stage} & \textbf{Server-GPU (s)} & \textbf{Desktop-CPU (s)} \\\hline
    \midrule
    \texttt{load} &  {\color{red} $0.883\pm 0.085$} & $0.786\pm 0.056$\\
    \texttt{fuse} & $0.306\pm 0.044$ & {\color{red} $3.003\pm 0.023$}\\
    \texttt{unfuse} & $0.206\pm 0.041$ & {\color{red} $2.916\pm 0.014$}\\
    \texttt{unload} & $0.007\pm 0.001$ & $0.007\pm 0.001$\\
    \bottomrule
\end{tabular}\vspace{-2mm}

    \caption{Latency (in s) to \texttt{load}, \texttt{fuse}, \texttt{unfuse}, \texttt{unload}~\cite{hflora} adapters on SDXL on Server-GPU and Desktop-CPU. On a mobile device, fusing/unfusing would happen for each layer iteratively since we cannot store all weights at the same time on local on-chip memory (unlike a large GPU), resulting in much higher overhead.}
    \label{tab:hflora}
\end{wraptable}
from Table~\ref{tab:hflora}, on a server-grade GPU, \texttt{load} time dominates whereas \texttt{fuse}/\texttt{unfuse}/\texttt{unload} times are relatively negligible. However, if we try to run the exact same pipeline on an everyday device like a desktop-grade CPU, we see that the \texttt{fuse} and \texttt{unfuse} times start dominating and can hinder rapid adapter switching. Note that, on an even more constrained device like a mobile phone, AI accelerators do not have sufficient memory to store weights from all layers at the same time in the local memory. Hence, on such devices, we would need to load base model weights for each layer into the local memory, and then fuse corresponding LoRA weights before we can run inference for that layer. This obviously leads to a massive inference latency overhead. As a result, none of the deployment options presented above are feasible for rapid adapter switching on mobile devices.

\section{Fuse and Scatter Op implementation}\label{sec::scatterop}
\begin{figure}[h]
    \centering
    \includegraphics[width=0.7\linewidth]{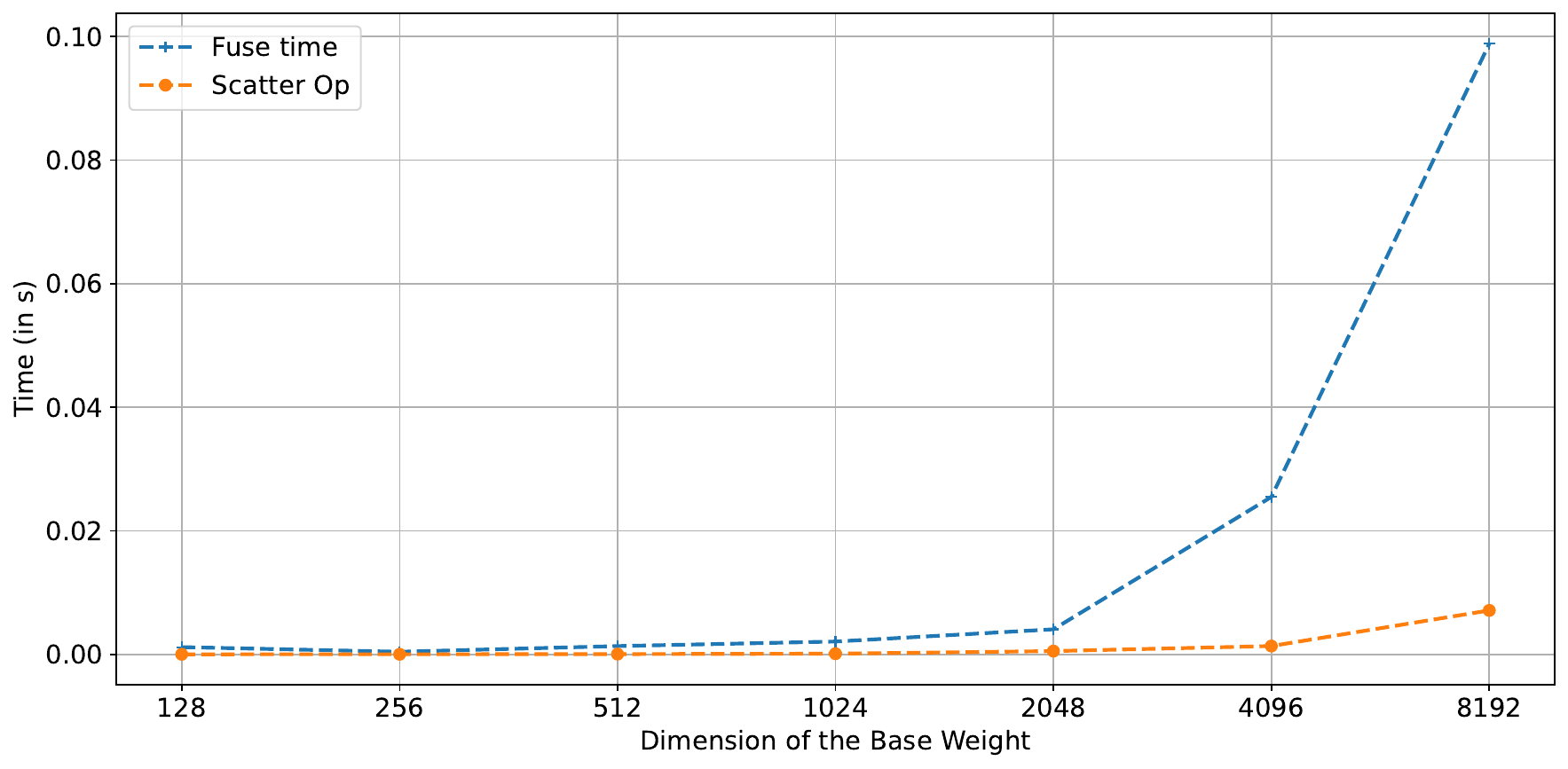}\vspace{-3mm}
    \caption{Comparison between average times for LoRA-fuse and SHiRA-\texttt{scatter\_op} implementation for 10 randomly initialized weights of various dimensions on a CPU (e.g., dimension = $4096$ means that the weight has shape $4096\times 4096$). For fusing, we compute time taken to merge LoRA adapters into the base weights (W + AB). Similarly, for the \texttt{scatter\_op}, we report time taken to overwrite base weights with SHiRA weights using the scatter op (\texttt{torch.Tensor.scatter\_}) based implementation in Pytorch.}
    \label{fig:speed}
\end{figure}

In this section, we compare fusing times of LoRA with our efficient \texttt{scatter\_op} (\texttt{torch.Tensor.scatter\_}) based implementation for SHiRA. For our experiments, we perform benchmarking on a Desktop-grade CPU and compute the average times for various tensor dimensions (e.g., tensor dimension = 4096 implies a weight of size $4096\times4096$, which is typical in modern LLMs). As shown in Fig.~\ref{fig:speed}, our \texttt{scatter\_op}-based SHiRA inference pipeline is up to $10\times$ faster than fusing LoRA weights, specially for larger dimensions.

\section{Turn any Trainer into SHiRA: Gradient Hook based Implementation}\label{Latencyeff}

In this section, we provide a method to adapt any floating point training into SHiRA based finetuning. Specifically, SHiRA can be implemented directly using a functionality called 
\texttt{post\_accumulate\_gradient\_hooks} available in Pytorch 2.1.0. This \texttt{gradient\_hook} can be used to mask gradients after the gradient accumulation step is completed. Moreover, this enables us to apply SHiRA on any publicly available trainer (e.g., \texttt{Transformers.Trainer, SFT\_Trainer}, etc.). Therefore, implementing SHiRA on any task is trivial and can be done even without PEFT library, thus making SHiRA very easy to implement.

With this gradient hook based implementation, we were able to train all our adapters (including for models such as LLaMA-7B, LLaMA2-7B and SD-1.5) on a single NVIDIA A100 GPU at nearly the same speed as PEFT based LoRA implementation. SHiRA runs at 2.17 it/sec as compared to LoRA which is at 2.42 it/sec for LLaMA-7B finetuning.

\section{Latency- and Memory-Efficient PEFT based Implementation for SHiRA}\label{Memoryeff}
As discussed in Appendix~\ref{sec::scatterop},  \texttt{scatter\_op}-based implementation can be utilized to manage sparse weight updates during inference. Given that SHiRA only finetunes a small subset of the pretrained model weights, we adopt a similar \texttt{scatter\_op}-based approach for training. This allows us to retain only the sparse training parameters in the optimizer, thereby significantly reducing the peak GPU memory utilization during training. As shown in Table~\ref{tab:memoryshira}, SHiRA not only trains at almost similar speed as LoRA, but also consumes $\sim16\%$ lower  peak GPU memory. Compared to other variants like DoRA, SHiRA training consumes significantly lower ($\sim40\%$) peak GPU memory and also trains much faster (SHiRA is about 36\% faster than DoRA). Finally, note that, partial finetuning techniques proposed in the pre-LoRA era do not have such memory-efficient implementations, which made them impractical for large generative models. All memory requirement data was collected using \texttt{psutil} utility used within the \texttt{Transformers.Trainer} training loop.
\begin{table}[h]
    \addtolength{\tabcolsep}{-2pt}
    \centering
    \fontsize{6.75pt}{6.75pt}\selectfont
    
    \begin{tabular} {  lccc }
    \toprule
    \centering \textbf{Adapter} & \textbf{Peak GPU memory (GB)} & \textbf{training steps/s} \\\hline
    \midrule
    \texttt{LoRA-PEFT} &  $35.10$ & $0.69$\\
    \texttt{DoRA-PEFT} & $49.49$ (\color{red} +40.99$\%$) & $0.49$ (\color{red} -28.98\%)\\
    \texttt{SHiRA-PEFT} & $29.26$ (\color{ForestGreen} -16.63\%) &  $0.67$ (\color{red} -2.89\%)\\
    \bottomrule
\end{tabular}\vspace{-2mm}

    \caption{Peak GPU memory consumption (in GBs) and Training steps per second during training for various PEFT-based implementation of adapters for LLaMA2-7B. Relative changes compared to LoRA are highlighted: {\color{ForestGreen} Green} indicates improved performance (lower memory consumption, faster training speed), while {\color{red} Red} indicates degraded performance (higher memory consumption, slower training speed).} 
    \label{tab:memoryshira}
\end{table}

\section{Dataset and Evaluation Metric Descriptions}\label{dataset-desp}
\subsection{Datasets}
\subsubsection{Language Datasets}
\begin{wraptable}{r}{5.0cm}
    \vspace{-0.7cm}
    \addtolength{\tabcolsep}{-2pt}
    \centering
    \fontsize{7.0pt}{5.75pt}\selectfont
    \begin{tabular} {  lccc }
    \toprule
    \textbf{Dataset} & \textbf{\#Train} & \textbf{\#Val} & \textbf{Test} \\\hline
    \midrule
    PiQA & 16K & 2K & 3K \\
    \midrule
    BoolQ & 9.4K & 2.4K &  2.4K \\
    \midrule
    SIQA & 33.4K & 1.9K &  1.9K \\
    \midrule
    OBQA & 4.9K & 0.5K &  0.5K\\
    \midrule
    Winogrande & 9.2K & 1.3K &  1.8K\\
    \midrule
    HellaSwag & 39.9K & 10K & 10K \\
    \midrule
    Arc\_easy & 2.25K & 570 & 2.36K \\
    \midrule
    Arc\_challenge & 1.12K & 299 & 1.12K \\
    \bottomrule
\end{tabular}
    \vspace{-2mm}
    \caption{  Commonsense Benchmark} 
    \label{tab: data_commonsense}
\vspace{-5mm}
\end{wraptable}
For language finetuning tasks, we use the commonsense reasoning datasets, which comprise 8 sub-tasks, each with a predefined training and testing set as shown in table \ref{tab: data_commonsense}. We follow the setting of \cite{llmadapters} for SHiRA Single Adapter training. The common sense reasoning training dataset is a combination of the training datasets provided by \cite{hudson2019gqa}, while we evaluate each evaluation dataset separately as in table \ref{tab: CommonSense-Single-Adapter}. For multi-adapter LLM experiments, we train each adapter from one particular task, and then perform multi-adapter evaluation on all the tasks.

\subsubsection{Vision Datasets}\label{visiondataset}
For style transfer adaptation tasks as described in sections~\ref{single-shira-lvm} and \ref{lvm-2}, we use two datasets, Bluefire and Paintings. Images present in both of these datasets are collected from public-domain (CC-0 license). 

The Bluefire dataset consists of a total of 54 images consisting of 6 different concepts - Cars, Dragons, Birds, Foxes, Men and Castles. For all these concepts, images with "blue-fire" effect are collected and used for style transfer finetuning. The validation of the Bluefire dataset consists of 30 images. 9 of the 30 images contain one of the 6 concepts in the training set, and the rest 21 are new. A few examples of unseen concepts in the validation set: \textit{football, monster, sword, chess rook, lion, koala etc}.

Similarly, the painting datasets contain a total of 90 images of "painting" style images of 9 different concepts - fire, birds, elephants, ships, horses, flowers, women, men and tigers. The validation set of the Paintings dataset consists of 21 images, out of which 9 contain concepts from the training set. The remaining 12 are new concepts not included in the training set. A few examples of unseen concepts in the validation set: \textit{lion, tiger, dog, cat, koala, panda, and other landscapes}.

\subsection{Evaluation Metrics} 
\paragraph{HPSv2 metric evaluation} 
For all style transfer finetuning experiments with Bluefire and Paintings dataset, we report HPS metric to quantify the quality of the generated images. For Bluefire validation, 30 images per validation prompt are generated for different seeds, hence generating 900 images for HPS analysis. We follow a similar paradigm for Paintings and generate 630 images with 21 prompts.

\section{Training Details}\label{training-details}
In this section, we list hyperparameters used for our experiments for Language and Vision finetuning tasks in table \ref{tab: train_details}.

\begin{table*}[h]
    \centering
    \fontsize{7.0pt}{5.75pt}\selectfont
    \begin{tabular} {  lcccccccc }
    \toprule
    
    \multirow{1}{*}{\textbf{Method}} & \multirow{1}{*}{\textbf{Adapter Target Modules}} & \multirow{1}{*}{\textbf{Optimizer}} &  \multirow{1}{*}{\textbf{LR}} & \multirow{1}{*}{\textbf{LR-Scheduler}} & \multirow{1}{*}{Rank}\\
    \hline
    \midrule

    LoRA LVM & \multirow{5}{*}{q-proj,k-proj,v-proj,up-proj,down-proj} & \multirow{5}{*}{AdamW} & $1e-4$ & Cosine & 64\\
    SHiRA LVM &  &  & $1e-4$ & Cosine & NA\\
    LoRA LLM &  &  & $2e-4$ & Linear & 32\\
    DoRA LLM &  &  & $2e-4$ & Linear & 32 \\
    SHiRA LLM & &  & $5e-4$ & Linear & NA \\

    \midrule
    \end{tabular}\vspace{-3mm}
    \caption{Training hyperparameters used for finetuning experiments.} 
\label{tab: train_details}
\end{table*}
\label{tab: detailsllm-multi}

 All finetuning and evaluation experiments for language and vision tasks are done using a single NVIDIA A100 GPU.

\section{Effect of Alpha}\label{alpha-effect}
\begin{figure}[h]
  \centering
   \includegraphics[width=1.0\linewidth]{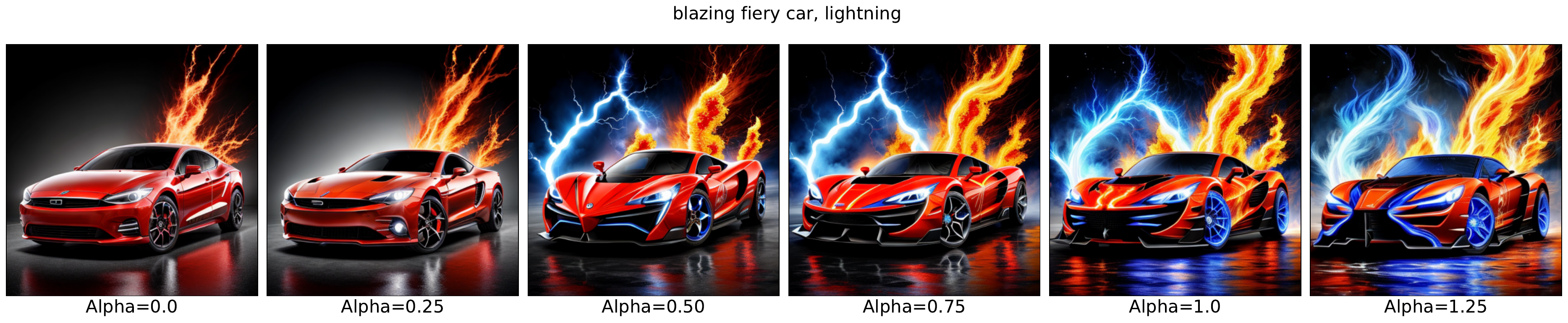}
   \vspace{-8mm}
   \caption{Effect of different $\alpha$ on SHiRA based Bluefire image generation.}
   \label{fig:alphaFigure}
\end{figure}

As described in the section~\ref{shira-approach}, in order to adapt the pretrained model to a new task, we only finetune very few weight parameters relevant to the task. For our adapter, we can easily extract out these modified weights as $S = W_{new} - W$, where $W_{new}$ is the weight obtained after SHiRA training, and $W$ is the prertained weight. Since only $1$-$2\%$ weights change during SHiRA training, $S$ is highly sparse and thus constitutes our sparse adapter. Hence, the new finetuned weights of the base model can be viewed as $W_{new} = W + S$.

Similar to LoRA, the strength of SHiRA adapter at inference time can be amplified using a scaling factor $\alpha$. For any defined $\alpha$ scaling, the new weights of the model can be expressed as $W_{new} = W + \alpha S$. Fig.~\ref{fig:alphaFigure} shows the effect of varying $\alpha$ on the output image. As evident, choosing an  $\alpha < 1$ reduces the "blue-fire" in the generated image and whereas $\alpha > 1$ amplifies the style transfer effect. For $\alpha=0.0$, the adapter is disabled and the model's output is the same as that for the base model.

\section{More Results}\label{AppMoreRes}
\begin{figure}[h]
  \centering
   \includegraphics[width=0.9\linewidth]{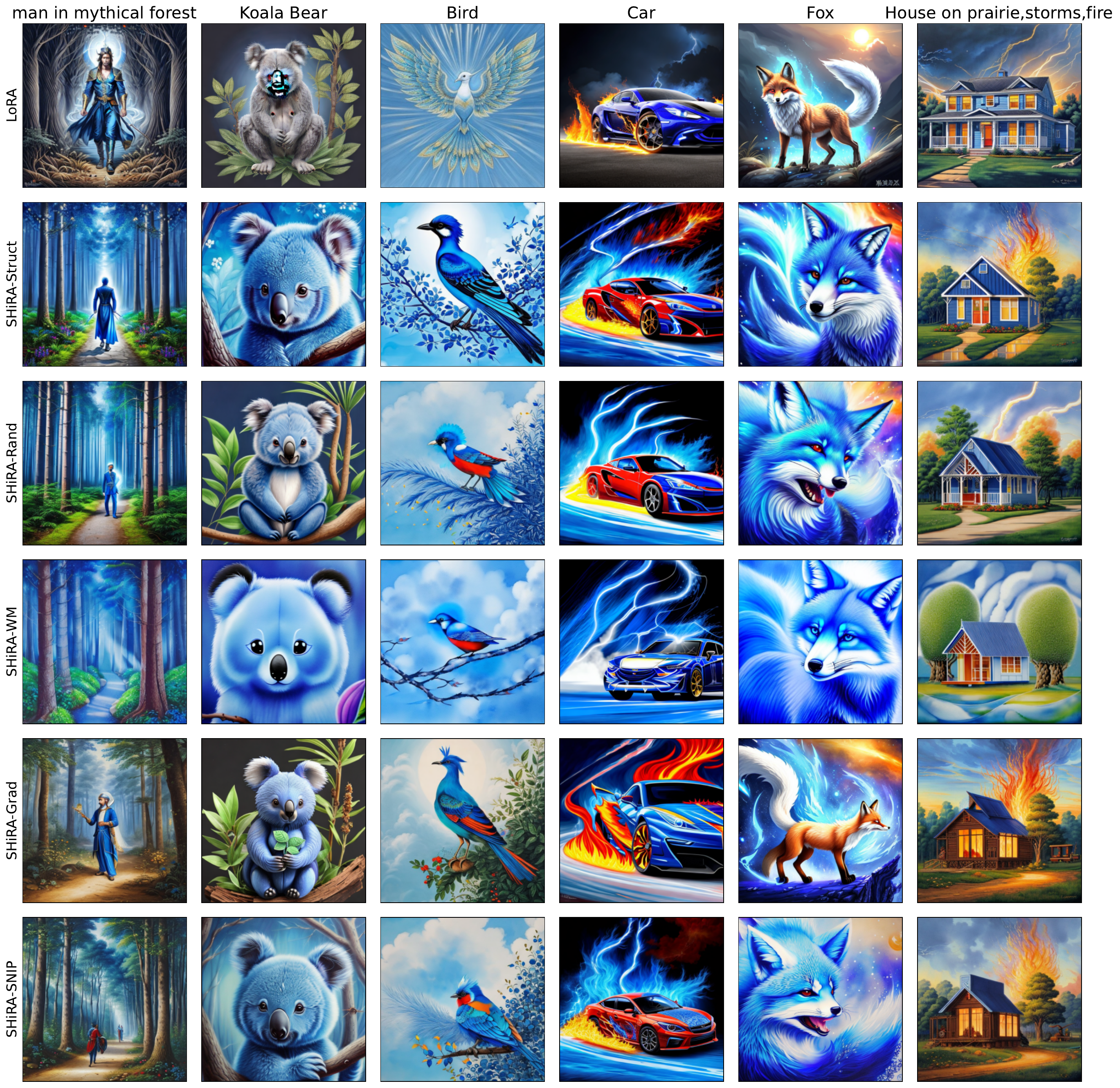}
   \caption{More results for multi-adapter fusion. Koala is not included in the training set of either of the Bluefire and Paintings Adapter styles. We observe that for this class, LoRA has significant artifacts whereas SHiRA produces exceptional images.}
   \label{fig:shiraFigure4}
\end{figure}

We show many more sample images for multi-adapter fusion in Fig.~\ref{fig:shiraFigure4}.

\clearpage
\label{sec:appendix}




\end{document}